\documentclass[fleqn,10pt]{wlscirep}
\usepackage[utf8]{inputenc}
\usepackage[T1]{fontenc}
\title{Haptic Shared Control Improves Neural Efficiency During Myoelectric Prosthesis Use}

\author[1,*]{Neha Thomas}
\author[2]{Alexandra J. Miller}
\author[3,4,5,6,7]{Hasan Ayaz}
\author[2]{Jeremy D. Brown}
\affil[1]{Department of Biomedical Engineering, Johns Hopkins University, Baltimore, 21218, USA}
\affil[2]{Department of Mechanical Engineering, Johns Hopkins University,  Baltimore, 21218, USA}
\affil[3]{School of Biomedical Engineering, Science and Health Systems, Drexel University, Philadelphia, PA, 19104 USA}
\affil[4]{Department of Psychological and Brain Sciences, Drexel University, Philadelphia, PA, 19104 USA}
\affil[5]{Drexel Solutions Institute, Drexel University, Philadelphia, PA, 19104 USA}
\affil[6]{Department of Family and Community Health, University of Pennsylvania, Philadelphia, PA, 19104 USA}
\affil[7]{Center for Injury Research and Prevention, Children's Hospital of Philadelphia, Philadelphia, PA, 19104  USA}
\affil[*]{neha.thomas@jhmi.edu}

\begin{abstract}
Clinical myoelectric prostheses lack the sensory feedback and sufficient dexterity required to complete activities of daily living efficiently and accurately. Providing haptic feedback of relevant environmental cues to the user or imbuing the prosthesis with autonomous control authority have been separately shown to improve prosthesis utility. Few studies, however, have investigated the effect of combining these two approaches in a shared control paradigm, and none have evaluated such an approach from the perspective of neural efficiency (the relationship between task performance and mental effort measured directly from the brain). In this work, we analyzed the neural efficiency of 30 non-amputee participants in a grasp-and-lift task of a brittle object. Here, a myoelectric prosthesis featuring vibrotactile feedback of grip force and autonomous control of grasping was compared with a standard myoelectric prosthesis with and without vibrotactile feedback. As a measure of mental effort, we captured the prefrontal cortex activity changes using functional near infrared spectroscopy during the experiment. Results showed that only the haptic shared control system enabled users to achieve high neural efficiency, and that vibrotactile feedback was important for grasping with the appropriate grip force. These results indicate that the haptic shared control system synergistically combines the benefits of haptic feedback and autonomous controllers, and is well-poised to inform such hybrid advancements in myoelectric prosthesis technology.
\end{abstract}
\begin{document}

\flushbottom
\maketitle
%
%
\thispagestyle{empty}


\section*{Introduction}

During volitional object manipulation, haptic sensations (proprioceptive, kinesthetic, and tactile) from the biological limb are used to make grasp corrections and update internal feedforward models of the object and environment \cite{Johansson1984RolesObjects}. This model refinement helps improve the speed and dexterity of subsequent manipulations, such that an initially hesitant interaction with an unknown or fragile object becomes smoother and more efficient with more experience \cite{Gordon1993MemoryObjects,Johansson1992Sensory-MotorActions}. Sensory information is particularly important for tuning grip forces to handle fragile or brittle objects; grip force must be great enough to counteract inertia and gravity, but not large enough to crush the object \cite{Gorniak2010ManipulationObject}. This haptic-informed knowledge is lost in typical upper-limb prostheses, as they do not provide sensory feedback. 

For the last several decades, researchers have been attempting to restore haptic feedback in upper-limb prostheses (see 2018 review by Stephens-Fripp, Alici, \& Mutlu  \cite{Stephens-Fripp2018}). In particular, significant effort has been placed on the use of mechanotactile stimulations on the skin to provide prosthesis wearers with cues like grip force, grip aperture, and object slip \cite{Witteveen2012,Antfolk2013a,Damian2018TheStability}. Prior research has demonstrated the benefits of haptic feedback in improving discriminative and dexterous task performance with a myoelectric prosthesis \cite{Thomas2019, Kim2012,Rosenbaum-Chou2016,Abd2022MultichannelDexterity}. Notably, vibrotactile feedback remains a simple, yet effective, method of haptic feedback in prostheses due to its compact size and low power consumption \cite{Raveh2018,Raveh2018AddingDisturbed,Stepp2012RepeatedPerformance,bark_comparison_2008,Blank2010,DAlonzo2012VibrotactileHand}. 

Despite the demonstrated benefits of haptic feedback for upper-limb prostheses, in particular for grip force modulation \cite{Kim2012,Rosenbaum-Chou2016, Clemente2016Non-InvasiveProstheses}, consistently controlling even standard myoelectric hands remains a challenge. In the simplest myoelectric scheme, direct control, the amount of electrical activity from an agonist-antagonist muscle pair is used to control a single degree-of-freedom prosthesis terminal device. The inherent delay between the user's interpretation of the haptic feedback and the subsequent myoelectric command could render volitional movement too slow \cite{Engeberg2011AdaptiveHands}, as well as cognitively demanding \cite{Jimenez2014,Thomas2020}. 

To reduce the user's cognitive burden while simultaneously improving task success, researchers have focused on embedding low-level autonomous intelligence directly on the prosthesis. These autonomous systems can react to and prevent grasping errors like object slip or excessive grasp force \cite{Salisbury1967APrehension, Chappell1987ControlHand, Nightingale1985MicroprocessorArm,Matulevich2013UtilityControl,Osborn2016}.  Similar techniques already enjoy commercial implementation, as is the case with the direct-control myoelectric Ottobock SensorHand Speed \cite{OttobockSensorHand} prosthesis. Beyond event-triggered autonomous systems, there are also controllers which attempt to optimize grasping performance, such as maximizing the contact area between the prosthetic hand and the object \cite{Zhuang2019SharedProsthesis}. Similarly, controllers that predict the likely sequence of desired prehensile grips without user intervention have been proposed \cite{Edwards2016ApplicationSwitching,Mouchoux2021ArtificialEffort,Mouchoux2021ImpactProstheses}. 

While these autonomous control strategies supplement human control to varying degrees, they fail to provide critical sensory feedback to the user during volitional prosthesis operation. This sensory feedback could be used to update the user's manipulation strategy and thereby improve volitional control. In turn, the autonomous controller could learn the human's successful volitional control strategies and replicate it in subsequent manipulations, thus improving task performance and reducing cognitive effort.

Control approaches that arbitrate between haptically-guided human control and autonomous control can be described as haptic shared control. Similar techniques have been incorporated in automotive applications, where haptic feedback from the autonomous system guides the driver during navigation \cite{Benloucif2019CooperativeDriving,Luo2021ADriving,Lazcano2021MPC-BasedDriving, Zhang2021HapticTeleoperation,Selvaggio2022ATransportation}.
In this manuscript, we explore the utility of a haptic shared control scheme for an upper-limb prosthesis. In particular, we investigate participants' ability to perform a dexterous grasp-and-lift task of a brittle object with either a standard myoelectric prosthesis, a myoelectric prosthesis with vibrotactile feedback of grip force, or a myoelectric prosthesis featuring haptic shared control --- that is, vibrotactile feedback of grip force and low-level autonomous control of grip force integrated through imitation-learning paradigm. Moreover, in addition to task performance, we also assess participants' cognitive load during task execution using functional near infrared spectroscopy (fNIRS), a noninvasive optical brain-imaging technique \cite{Ayaz2012OpticalAssessment}. This experiment builds from two previous investigations where we separately evaluated the cognitive load associated with haptic feedback during prosthesis use \cite{Thomas2020} and the task performance utility of a haptic shared control scheme \cite{Thomas2021Sensorimotor-inspiredVision}  

Here, we employ the same neurophysiological measurements to provide a holistic understanding of performance and mental effort required to achieve that performance, i.e., neural efficiency \cite{Curtin2018}. We hypothesize that haptic shared control will result in the highest neural efficiency compared to the standard prosthesis, followed by the prosthesis featuring vibrotactile feedback of grip force. 
\section*{Methods}

\subsection*{Participants}
\begin{figure}
\centering
\includegraphics[width=.3\textwidth]{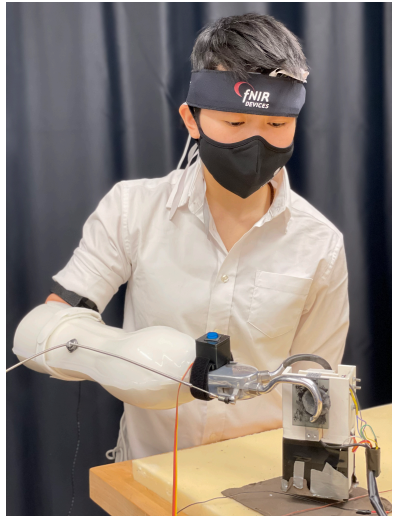} 
\caption{User grasps the brittle instrumented object with a myoelectric prosthesis featuring haptic shared control. An fNIRS headset over the forehead gathers neurophysiological measures of cognitive load. }
\label{fig: person}
\end{figure}

33 non-amputee participants (9 female, age 24.6 $\pm$ 3.20, 2 left-handed individuals) participated in this experimental study approved by the Johns Hopkins Medical Institute IRB (protocol\# 00147458). Participants were pseudo-randomly assigned to one of three groups, and each group was balanced for gender. Participants in the first group completed a grasp-and-lift task using a standard myoelectric prosthesis (Standard group). Participants in the second group completed the same grasp-and-lift task using a myoelectric prosthesis with vibrotactile feedback of grip force (Vibrotactile group). Participants in the third group completed the grasp-and-lift task using a myoelectric prosthesis featuring haptic shared control (Haptic Shared Control group). Fig. \ref{fig: person} shows one of the experimenters demonstrates the experimental setup for the Haptic Shared Control Group. All participants provided informed consent before participating. 

\subsection*{Experimental Hardware}
The devices used in the experiment include a mock prosthesis, vibrotactile actuator, fNIRS imaging device, and an instrumented object. Excluding the fNIRS data stream, all input and output signals were controlled through a Quanser QPIDe DAQ and QUARC real-time software in MATLAB/Simulink 2017a. 

\subsubsection*{Prosthesis}
The prosthesis consists of a custom thermoplastic socket that can be worn by non-amputee participants and a voluntary-closing hook-style terminal device (maximum aperture 83\,mm). A Bowden cable connects the terminal device to a custom motorized linear actuator to control device opening and closing. This same prosthesis and actuator have been used and described in more detail in Thomas et al. \cite{Thomas2019,Thomas2020}. A counterweight system was attached to the terminal device to simulate the loading conditions typically experienced by transradial amputees; it offset 500\,g of the prosthesis's 800\,g mass. 

The motorized linear actuator is driven in proportional, open-loop speed control mode by surface electromyography signals (sEMG) from the wrist flexor and extensor muscle groups. sEMG signals were acquired using a 16-channel Delsys Bagnoli Desktop system. 

\subsubsection*{Instrumented Object}

\begin{figure}
\centering
\includegraphics[width=.4\textwidth]{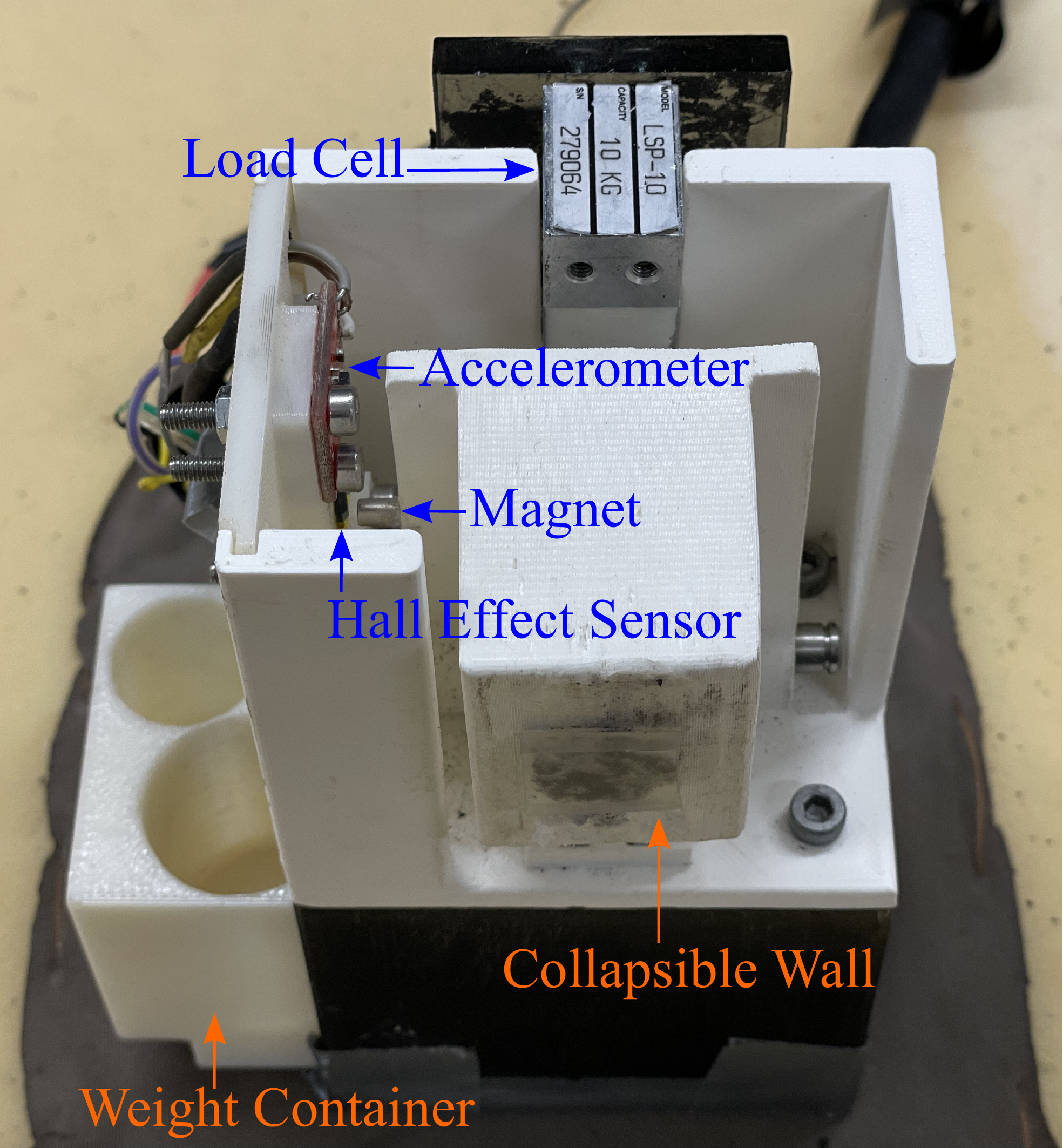} \caption{The instrumented object simulates breaks using a collapsible wall and is capable of measuring grip force and object movements with a load cell and accelerometer.}\label{fig: test object}
\end{figure}

Inspired by previous research \cite{Meek1989a, Brown2015a}, an instrumented device that simulates a brittle object ($77 \times 74 \times 139\,\textrm{mm}$) was designed for the grasp-and-lift task. This object, depicted in Fig. \ref{fig: test object}, consists of a collapsible wall to signify object breakage. The object features an accelerometer to measure object movements, a magnet and Hall effect sensor to detect breaks, a 10\,kg load cell to measure grip force, and a weight container to customize the object's mass. For the present study, the mass of the object remained constant at 310\,g. Conductive fabric was placed on the base of the object and surface of the testing platform to detect object lift.

\subsubsection*{Non-Invasive Brain Imaging}
Functional near infrared spectroscopy (fNIRS) operates on the principles of relating neuronal activity to metabolic activity \cite{Ayaz2012OpticalAssessment,Kuschinsky1991CouplingBrain}. It has been shown to produce similar results as the brain imaging gold standard, fMRI \cite{Fishburn2014SensitivityLoad,speakerlistenerfMRIfNIRS}, and has been demonstrated in numerous prior studies \cite{Ayaz2013ContinuousDevelopment.,Perrey2014PossibilitiesFNIRS,Mirelman2014IncreasedAdults,Gateau2018}. Furthermore, we have previously used fNIRS to successfully assess the effect of a myoelectric prosthesis featuring vibrotactile feedback on cognitive load in a stiffness discrimination task \cite{Thomas2020}. 
A four-optode fNIRS imager (Model 1100W; fNIR Devices, LLC) was used to measure hemodynamic activity from four regions of the prefrontal cortex. 
Signals were acquired and post-processed in COBI Studio (v1.5.0.51) \cite{Ayaz2011UsingNavigation}. The raw light intensity data is first filtered using a 40 order low-pass and linear-phase Finite Impulse Response (FIR) filter using a Hamming window and  cut-off frequency of 0.1Hz to attenuate high frequency noise and physiological oscillations like heart rate and respiration rate. Then the modified Beer Lambert is applied to the filtered data to obtain the relative concentrations of hemoglobin,  which are related to mental effort via neurovascular coupling \cite{Ayaz2012OpticalAssessment,Kuschinsky1991CouplingBrain}.

\subsection*{sEMG Calibration and Prosthesis Manual Control}
\label{sec: Prosthesis Control}
The sEMG calibration procedure employs maximum voluntary contraction to normalize the sEMG signals between the minimum and maximum voltages of the prosthesis motor. The amplitude of wrist flexion activity is proportional to the closing speed of the prosthesis. Likewise, the amplitude of wrist extension activity is proportional to the opening speed of the prosthesis. The control equation and more details can be found in our previous study \cite{Thomas2021Sensorimotor-inspiredVision}; the only difference here is the range of voltages used to drive the current prosthesis (1.5 -- 7\,V here instead of 0.55 -- 1.5\,V in our prior work).

\subsection*{Vibrotactile Feedback}
\label{sec: Vib}
Vibrotactile feedback of grip force was provided using a C-2 tactor (Engineering Acoustics) driven by a Syntacts amplifier (v3.1) \cite{Pezent2021Syntacts:Haptics}. The tactor was strapped to the upper-arm of the participant.
Vibrotactile feedback frequency was set to 250\,Hz. Vibrotactile feedback voltage $\nu$ was proportional to the load cell voltage $L$ from the instrumented object. As the force on the load cell increased, the amplitude of the vibration increased as shown in 
\begin{equation}
    \nu = 10 \cdot \frac{4.3\,\textrm{V}-L}{4.3\,\textrm{V}} \cdot \sin({2 \pi \frac{\textrm{rad}}{\textrm{cycle}} \cdot 250\,\text{Hz}} \cdot t)
\end{equation}

The resting state of the load cell is around 4.5\,V. As force is applied to the load cell, this value decreases. 4.3\,V was chosen as the threshold for detecting contact on the load cell.

\subsection*{Haptic Shared Controller}
\label{sec: shared control}
The haptic shared control scheme switches between the user's manual control (with vibrotactile feedback) of the prosthesis and an autonomous control system that attempts to mimic the user's desired grip force. When enabled and subsequently triggered by the user, the autonomous controller independently closes the prosthesis terminal device until the user's preset grip force is achieved. 

In order to enable the controller, the user must first manually actuate the prosthesis (via sEMG) and lift the object for a minimum of one second without breaking or dropping the object. Such occurrences are identified by assessing sharp peaks in the derivative of the load cell signal, where $\frac{dL}{dt} > 2.5 \frac{\textrm{V}}{\textrm{s}}$ indicates a slip event, and $\frac{dL}{dt} > 5 \frac{\textrm{V}}{\textrm{s}}$ indicates object breakage.
During this manual operation, the participant receives vibrotactile feedback of the grip force as described in the Vibrotactile Feedback section. The average applied grip force during the one second of successful lifting is stored as the desired grip force for the shared controller. Once this value is stored, the vibrotactile feedback turns off and the blue LED on the prosthesis turns on, informing the user of the transition to autonomous control.

To activate the autonomous closing of the prosthesis terminal device, participants have to generate a wrist flexion sEMG signal $S_f$ greater than or equal to the wrist flexion threshold, $f_L$. Closing of the terminal device occurred in three separate stages. First, an initial decaying signal initiated fast closing as shown in

\begin{equation}
h^1_c = 2.5 \cdot \textrm{max}[0.3, e^{-1t}] ,  \ \ \ S_f > f_{L} \\
\label{eqn: stage 1}
\end{equation}

Once the closing speed slowed below a heuristically determined threshold, the closing command was ramped up continuously until contact with the object was detected as in 

\begin{equation}
h^2_c = \textrm{max}[-0.5, \textrm{min}(e^{t},4)] , \ \ \ a_L < \frac{da}{dt} < a_U \\
\label{eqn: stage 2}
\end{equation}
\noindent where $\frac{da}{dt}$ is the derivative of the prosthesis aperture (i.e., velocity), $a_L$ refers to the lower velocity threshold, and $a_U$ is the upper threshold. 

Contact occurred when the load cell value $L$ on the brittle object decreased below a threshold $L_t$ and the aperture of the prosthesis $A$ decreased below a threshold $A_t$, measured by an encoder on the motor of the linear actuator. After contact, a proportional and integral controller closes the terminal device until the load cell signal is within 5\% of the user's predefined grip force $L_{\textrm{d}}$ as shown in 

\begin{equation}
 h^3_c = \textrm{max}[0,\textrm{min}[ (L_d-L) + 3\int_{0}^{t} (L_d-L),12], \ \ \ L<0.95 L_d \ \& \ A < A_t\\ 
 \label{eqn: stage 3}
\end{equation}

    
 
\begin{figure}
\centering
\includegraphics[width=.5\textwidth]{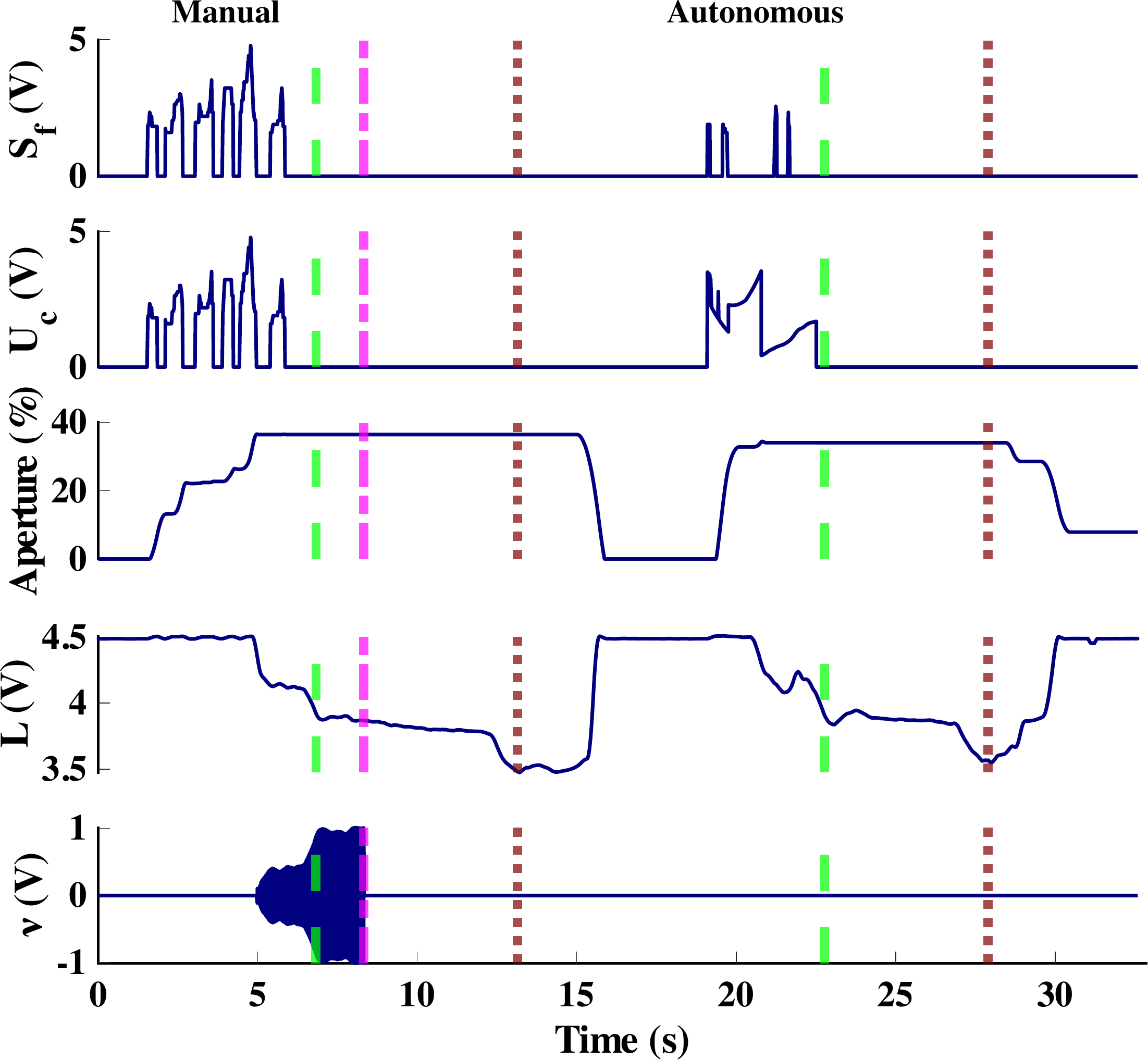} \caption{Example signals from the first trial of a participant in the Haptic Shared Control condition as they grasped and lifted the brittle object. The green dashed lines indicate when the object was lifted, and the brown dotted lines indicate when the object was set down. The pink dashed lines indicate when the autonomous controller was enabled. The traces shown are sEMG flexion activity $S_f$, closing command $u_c$, the percent closed of the prosthesis, the load cell signal $L$, and the C-2 tactor vibration signal $\nu$. The traces depict two successful grasp-and-lifts, where the first attempt was done manually with vibrotactile feedback, while the second attempt was completed using the autonomous control.}
\label{fig: traces}
\end{figure}

The autonomous controller is disabled if the object breaks or is dropped. The controller can also be manually disabled by the user by pushing the blue LED button on the prosthesis. When the controller is disabled, the user receives a short, pulsed vibration and the LED turns off. Signal traces for a participant using the haptic shared controller can be seen in Fig. \ref{fig: traces}. 

\subsection*{Experiment Procedure}

Before starting the experiment, each participant completed a demographics questionnaire. Next, an experimenter placed an sEMG electrode on the participant's right wrist flexor muscle group and another on their right wrist extensor muscle group. Participants calibrated their sEMG signals using maximum voluntary contraction (MVC) of their wrist flexor and extensor muscle groups 
Experimenters then placed the fNIRS imaging headset on the participant's forehead and took baseline measurements of prefrontal cortex activity. 

While seated in front of the experimental table, participants used a GUI to complete a combined training and assessment of their sEMG signal control, modeled after the test reported in \cite{Prahm2018PlayBionic:Rehabilitation}. Participants were asked to reach and sustain three different levels of sEMG activity for five seconds at a time; each of the levels were 12.5\%, 25\%, and 37.5\% of the user's MVC. The participant first completed one practice session for wrist flexor activity, where each of the three levels was presented once. Once practice was complete, the participant completed a test session, in which each of the three levels was presented three times. After completing the practice and testing sessions for the wrist flexor,  participants repeated the same practice and test procedure for the wrist extensor. 

After completing the sEMG training and assessment, participants were asked to stand to begin training for the grasp-and-lift task. If the participant was in the Vibrotactile or Haptic Shared Control group, the C-2 tactor was placed on their upper right arm. Likewise, if the participant was in the Haptic Shared Control group, the blue LED button was placed on the prosthesis. The experimenter then instructed the participant on how to close and open the prosthesis using their muscle activity. Participants were able to practice closing and opening the prosthesis until they felt comfortable. Next, the experimenter explained that the goal of the task was to grasp and lift the brittle instrumented object for three seconds without breaking or dropping it. 

Participants in the Standard group were allowed multiple attempts until they successfully lifted the object for three seconds. They were then given three more practice attempts before moving on to the actual experiment. 

Participants in the Vibrotactile group were first given an overview of the feedback and were further instructed to use the feedback to find the appropriate grip force for lifting the object. They were then allowed multiple attempts until they could successfully lift the object for three seconds. Afterwards, they were given three more practice attempts before moving on to the actual experiment.

Participants in the Haptic Shared Control group were first given an overview of the shared controller, and informed about how to switch between manual and autonomous modes. They were then allowed multiple attempts until they could successfully lift the object for three seconds in the manual mode. Next, they were asked to trigger the autonomous control and lift the object (see the Haptic Shared Controller section). The experimenter then demonstrated the two scenarios that automatically disabled the autonomous controller: (1) an object break, and (2) an object slip. The participant was then required to grasp and lift the object in manual mode after each demonstration in order to re-enable the autonomous controller. Finally, the experimenter demonstrated how to use the blue LED button to manually override the autonomous controller. Afterwards, the participant was allowed two more practice attempts before moving on to the actual experiment. Participants in the Haptic Shared Control group began the experiment in manual mode.

After all training had been completed, participants then completed seven one-minute trials of the grasp-and-lift task, wherein they attempted to grasp and lift the object as many times as possible within that minute without breaking or dropping the object. Participants were required to hold the object in the air for 3 seconds. A 30-second break was provided between trials.

After finishing all seven trials, participants then completed a survey regarding their subjective experience of the experiment. The questions were based on the NASA-TLX questionnaire \cite{Hart1988DevelopmentResearch} and included a mix of sliding scale and short answer questions. 

\subsection*{Metrics}
The following metrics were used to analyze the three conditions from the perspective of both task performance and neural performance. 

\subsubsection*{Task Performance}
A successful lift was defined as lifting and holding the object in the air for at least three seconds. There were no requirements for lifting height in the task. The status of each grasp attempt (successful lift or not) was recorded. In addition, the total number of successful lifts per trial was also calculated.

A safe grasping margin was defined for the instrumented object as a load cell value in the range of 3-4\,V. For each grasp attempt, the 100 smallest load cell values (measured during object grasp, representing the maximum force values -- refer to Fig. \ref{fig: traces}) were averaged and compared to the safe grasping interval.

\subsubsection*{Neural Performance}
The total concentration of hemoglobin (HbT) was used as a proxy for measuring cognitive load. The average value was extracted for each of the seven trials from four regions of the prefrontal cortex: left lateral, left medial, right medial, and right lateral.

These cognitive load measurements were combined with the total number of lifts to calculate neural efficiency as described in \cite{Curtin2018}. The z-scores of the number of lifts $z(\textrm{Lift})$ and total hemoglobin concentration $z(\textrm{HbT})$ were calculated to derive the neural efficiency metric as

\begin{equation}
    N = \frac{z(\textrm{Lifts})-z(\textrm{HbT})}{\sqrt{2}} 
\label{eqn: neural efficiency}    
\end{equation}

This metric describes the mental effort required to achieve a certain level of performance. A higher neural efficiency is associated with higher performance and lower cognitive load.

\subsubsection*{Survey}
The post-experiment survey was a mix of sliding (0-100) and short answer questions. The sliding scale questions asked participants to rate the task's physical demand, mental demand, and pacing. Additionally, it asked them to rate their perceived ability to complete the task, their frustration level and how much they used visual, auditory, and touch-based cues to help them complete the task. Finally, the survey prompted them to explain the strategy they employed to accomplish the task and provide any other comments about their experience.

\subsection*{Statistical Analysis}

Statistical analysis was carried out in RStudio (v4.1.0). A mixture of logistic and linear mixed models were used to assess task and neural performance. The random effects included a random intercept for the subject and a random slope for trial. Post-hoc tests were conducted with a Bonferroni correction. Model residuals were plotted and checked for homogeneity of variance and normality. A p-value of 0.05 was used as the threshold for significance.

A logistic binomial mixed model was used to assess the probability of being within the safe grasping margin for each grasp attempt. The fixed effects included the trial number and the mode, where mode could be No Feedback (manual operation of the myoelectric prosthesis), Feedback (manual operation of the prosthesis with vibrotactile feedback of grip force), or Autonomous (autonomous controller operates prosthesis). The Feedback grasp attempts includes all grasp attempts from participants in the Vibrotactile group and grasp attempts from participants the Haptic Shared Control group who were manually operating the prosthesis (autonomous controller disengaged).  

A separate logistic binomial mixed model was used to assess the probability of lifting the object. Individual linear mixed models were used to assess the number of lifts, the total concentration of hemoglobin for each of the four brain regions, and the neural efficiency for each of the four brain regions. The fixed effects for all models were the participant group and the trial number. For this analysis, the Vibrotactile group is separate from the Haptic Shared Control group, and does not include trials from participants in the Haptic Shared Control group manually operating the prosthesis.

\section*{Results}

Three of the thirty-three participants who consented to participate in the study were excluded from data analysis. Of those three, one was unable to finish the experiment due to technical issues with the system. Another participant was unable to produce satisfactory sEMG signals during the calibration step, and a third participant had poor control during the experiment. This participant also exhibited poor control during the sEMG assessment, as evidenced by their high root-mean-square error during the flexion and extension sEMG assessment compared to the other participants. The following results are for the remaining 30 participants (10 in each group). 

\subsection*{Outcome Measures}

Results reported for the data indicate the estimate of the fixed effects $\beta$ and the standard error $SE$ from linear and logistic mixed models statistical analyses. For brevity, cognitive load results from only the right lateral prefrontal cortex will be discussed, as this region presented the most significant changes in activity. Cognitive load results from the other brain regions can be found in the supplemental material associated with this manuscript. 

\subsection*{Task Performance}

\begin{figure}
\centering
\includegraphics[width=.4\textwidth]{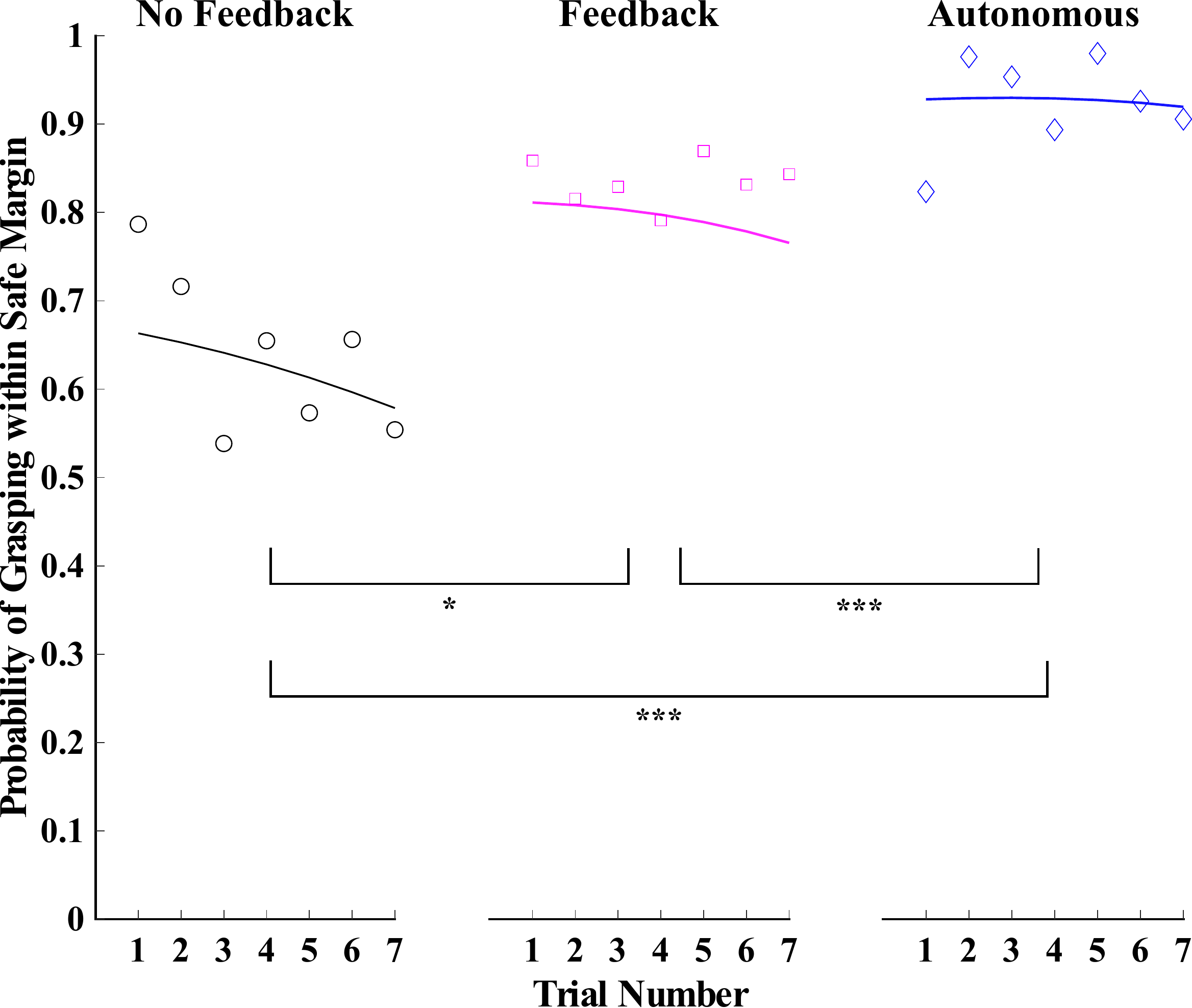} \caption{The probability of grasping the object within a safe grip force margin for each grasp attempt, where the individual data points represent the average for each trial (for all participants in each condition), and the solid lines indicate the model's prediction. Note: the No Feedback mode refers to manual prosthesis operation, the Feedback mode refers to manual prosthesis operation with vibrotactile feedback, and the Autonomous mode refers to autonomous prosthesis operation. * indicates $p<0.05$, ** indicates $p<0.01$, and *** indicates $p<0.001$.}
\label{fig: safe margin vib}

\end{figure}

\subsubsection*{Safe Grasping Margin}
A binomial mixed model was used to assess the odds that a given grasp attempt was adequate for lifting the object without breaking it. Here, we compare the No Feedback, Feedback, and Autonomous modes. The No Feedback mode includes all participants in the Standard group. The Feedback mode includes participants in the Vibrotactile group as well as the Haptic Shared Control participants who were in the manual mode. The Autonomous mode includes Haptic Shared Control participants using the autonomous controller to complete the grasp-and-lift task. The odds of being within a safe grasping margin were approaching a significant positive difference from 50\% in the Standard mode ($\beta=0.96, SE=0.51, p=0.06$). However, both the Vibration mode ($\beta=1.07, SE=0.53, p=0.045$) and the Autonomous mode ($\beta=2.61, SE=0.57, p<0.001$) significantly improved the odds of being withing a safe grasping margin compared to the Standard mode. Furthermore, the Autonomous mode was significantly better than the Vibrotactile mode ($\beta=1.55, SE=0.33, p<0.001$) in ensuring a safe grasping margin. Experience with the task (i.e., number of trials) had no effect on the ability to grasp within the safety margin ($\beta=-0.07, SE=0.06, p=0.22$). See Fig. \ref{fig: safe margin vib} for a visualization of these results.

\subsubsection*{Lifting Probability}
\begin{figure}[b!]
\centering
\includegraphics[width=.4\textwidth]{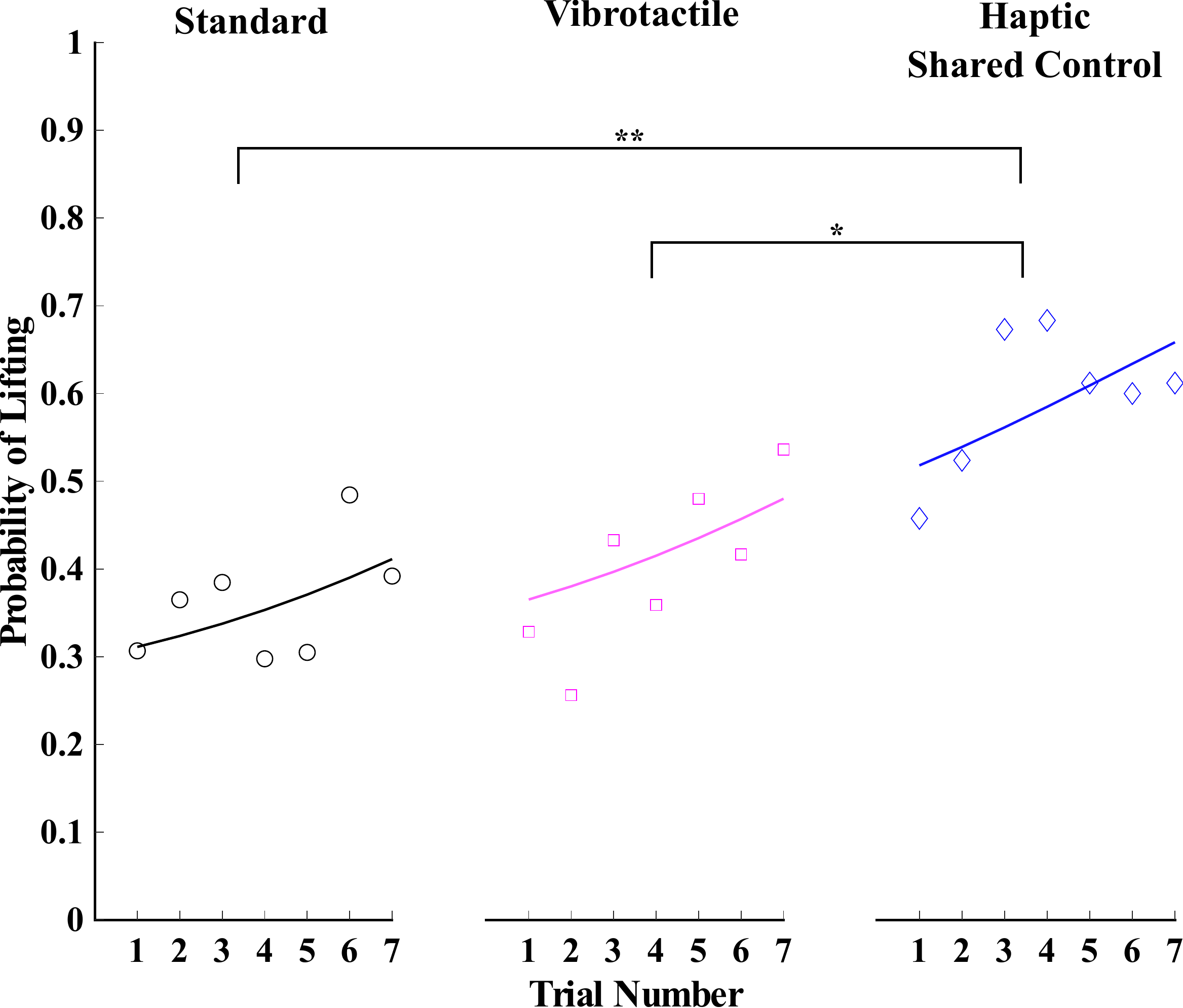} \caption{The probability of lifting the object for each group across trials, where the individual data points represent the average for each trial (for all participants in each group), and the solid lines indicate the model's prediction. * indicates $p<0.05$, ** indicates $p<0.01$, and *** indicates $p<0.001$.}
\label{fig: prob lift}
\end{figure}

A binomial mixed model was used to assess the odds that a given grasp attempt resulted in a successful lift. Here and in all subsequent results, we compare the Standard, Vibrotactile and Haptic Shared Control groups.
The odds of lifting the object in the Standard group were significantly less than 50\% ($\beta=-0.96, SE=0.35, p=0.006$). The Vibrotactile group was not better than the Standard group ($\beta=0.30, SE=0.37, p=0.42$). However, the Haptic Shared Control group significantly improved the probability of lifting the object compared to the Standard group ($\beta=1.08, SE=0.36, p=0.003$) and the Vibrotactile group ($\beta=0.78, SE=0.37, p=0.037$). In addition, experience with the task (i.e., number of trials) significantly improved performance across all groups ($\beta=0.09, SE=0.04, p=0.025$). See Fig. \ref{fig: prob lift} for a visualization of these results for each group.

\begin{figure}
\centering
\includegraphics[width=.4\textwidth]{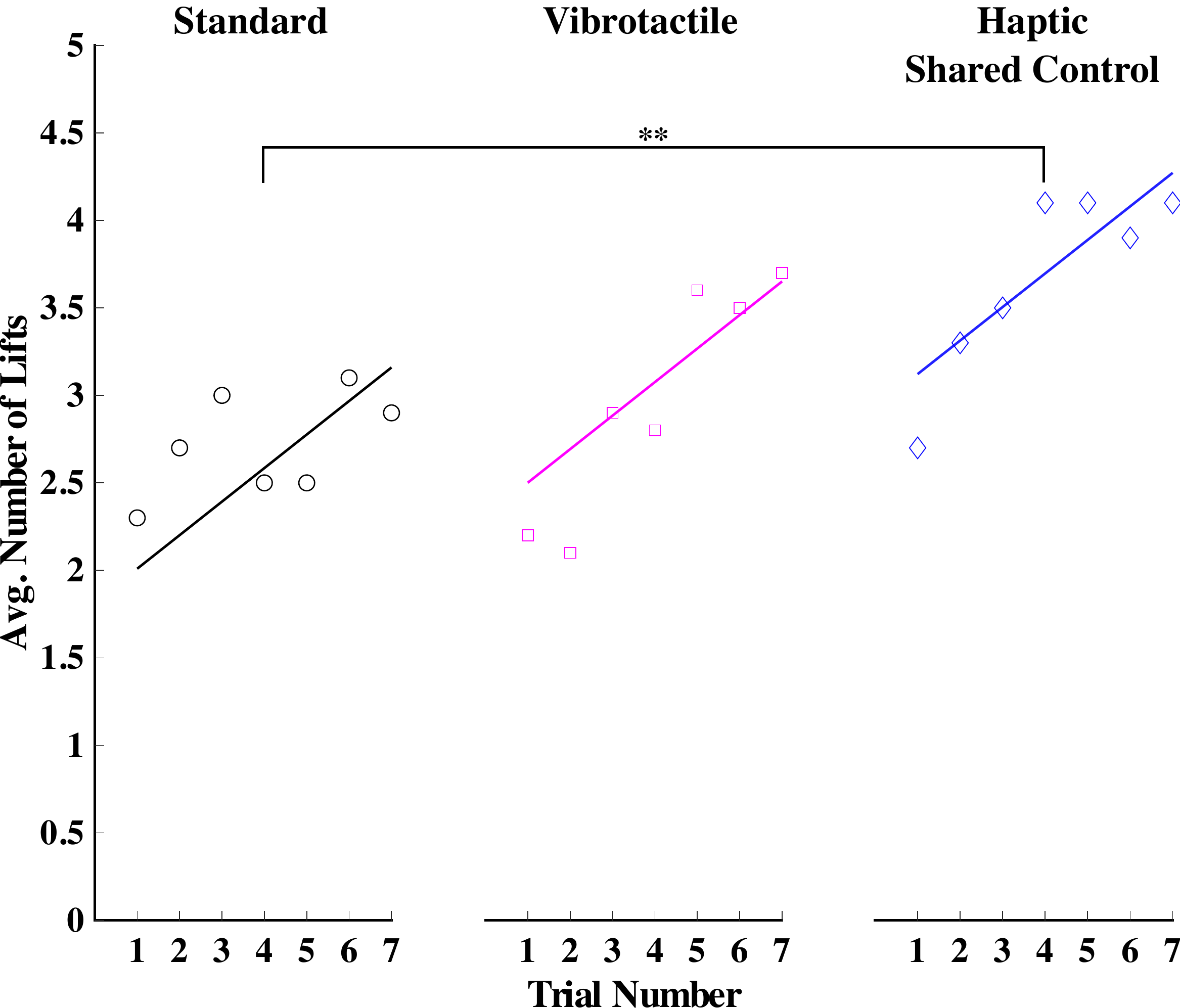} \caption{The average number of lifts for each group across trials, where the individual data points represent the average for each trial (for all participants in each group), and the solid lines indicate the model's prediction. * indicates $p<0.05$, ** indicates $p<0.01$, and *** indicates $p<0.001$.}
\label{fig: nlift}
\end{figure}

\subsubsection*{Number of Lifts}
A linear mixed model was used to assess the average number of three-second lifts per trial.
The average number of lifts in the Standard group was significantly higher than zero ($\beta=1.82, SE=0.42, p<0.001$). The Vibrotactile group was not different from the Standard group ($\beta=0.50, SE=0.51, p=0.34$) or the Haptic Shared Control group ($\beta=--0.62, SE=0.51, p=0.22$). However, the Haptic Shared Control group significantly improved the number of lifts compared to the Standard group ($\beta=0.96, SE=0.35, p=0.006$). In addition, experience with the task significantly improved performance across all groups ($\beta=0.19, SE=0.04, p<0.001$). See Fig. \ref{fig: nlift} for a visualization of these results for each group.

\begin{figure}
\centering
\includegraphics[width=.4\textwidth]{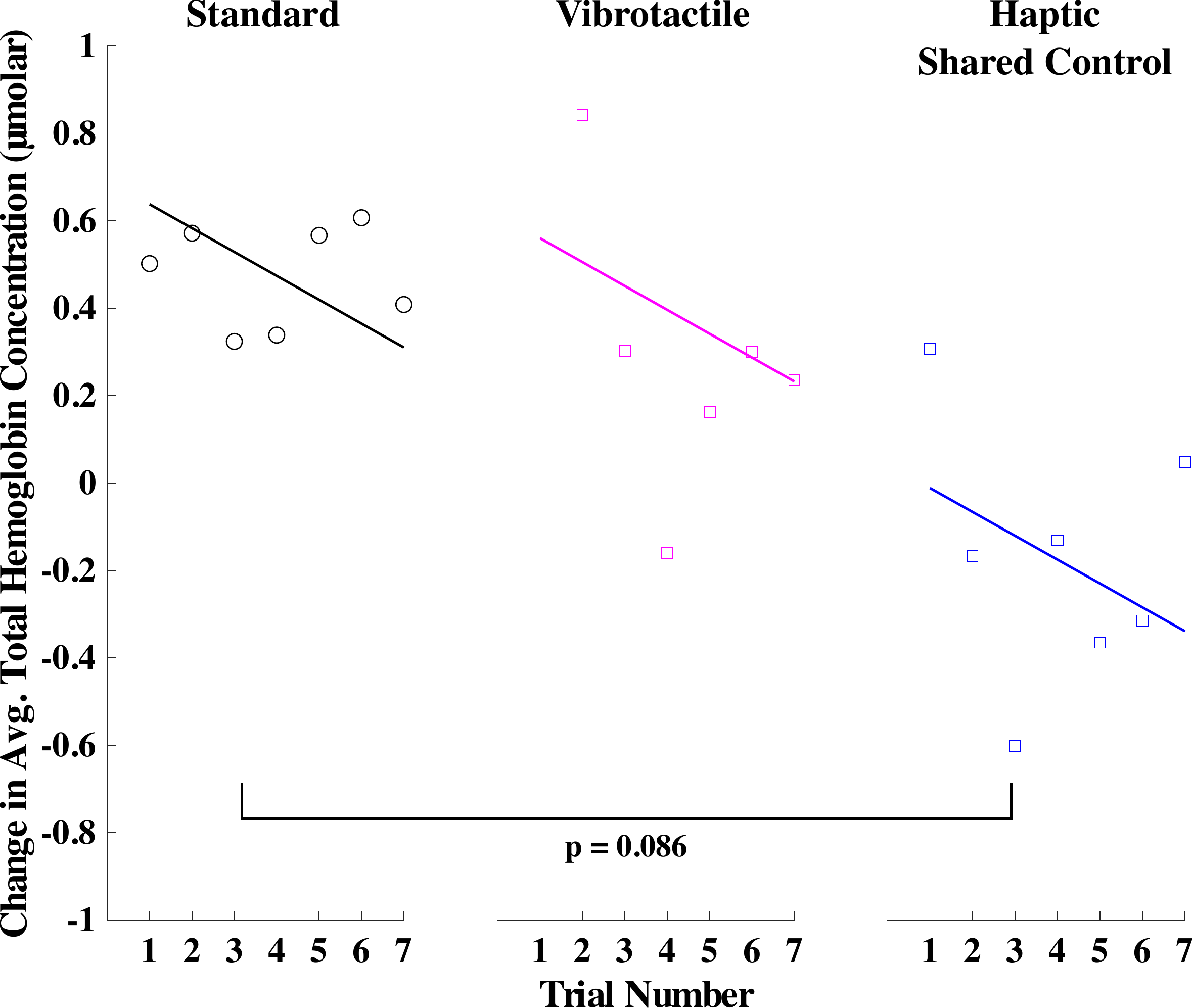} 
\caption{The average total hemoglobin concentration for each trial, where the individual data points represent the average for each trial (for all participants in each group), and the solid lines indicate the model's prediction. * indicates $p<0.05$, ** indicates $p<0.01$, and *** indicates $p<0.001$.}
\label{fig: HbT4}
\end{figure}


\subsection*{Neural Performance}

\subsubsection*{Change in Average Total Hemoglobin Concentration}
The change in average total hemoglobin concentration represents the amount of cognitive load incurred. An increased concentration indicates a higher cognitive load. A linear mixed model was used to assess the hemoglobin concentration. The average total hemoglobin concentration in the right lateral prefrontal cortex was significantly higher than zero in the Standard group ($\beta=0.71, SE=0.29, p=0.019$). The Vibrotactile group was not significantly different from the Standard group ($\beta=-0.10, SE=0.37, p=0.77$). Similarly, the  Haptic Shared Control group was not significantly different from the Standard group ($\beta=-0.65, SE=0.37, p=0.086$). Experience with the task was close to significantly improving the cognitive load (reducing total hemoglobin concentration: $\beta=-0.05, SE=0.03, p=0.068$). See Fig. \ref{fig: HbT4} for a visualization of these results for each group.

\begin{figure}
\centering
\includegraphics[width=.4\textwidth]{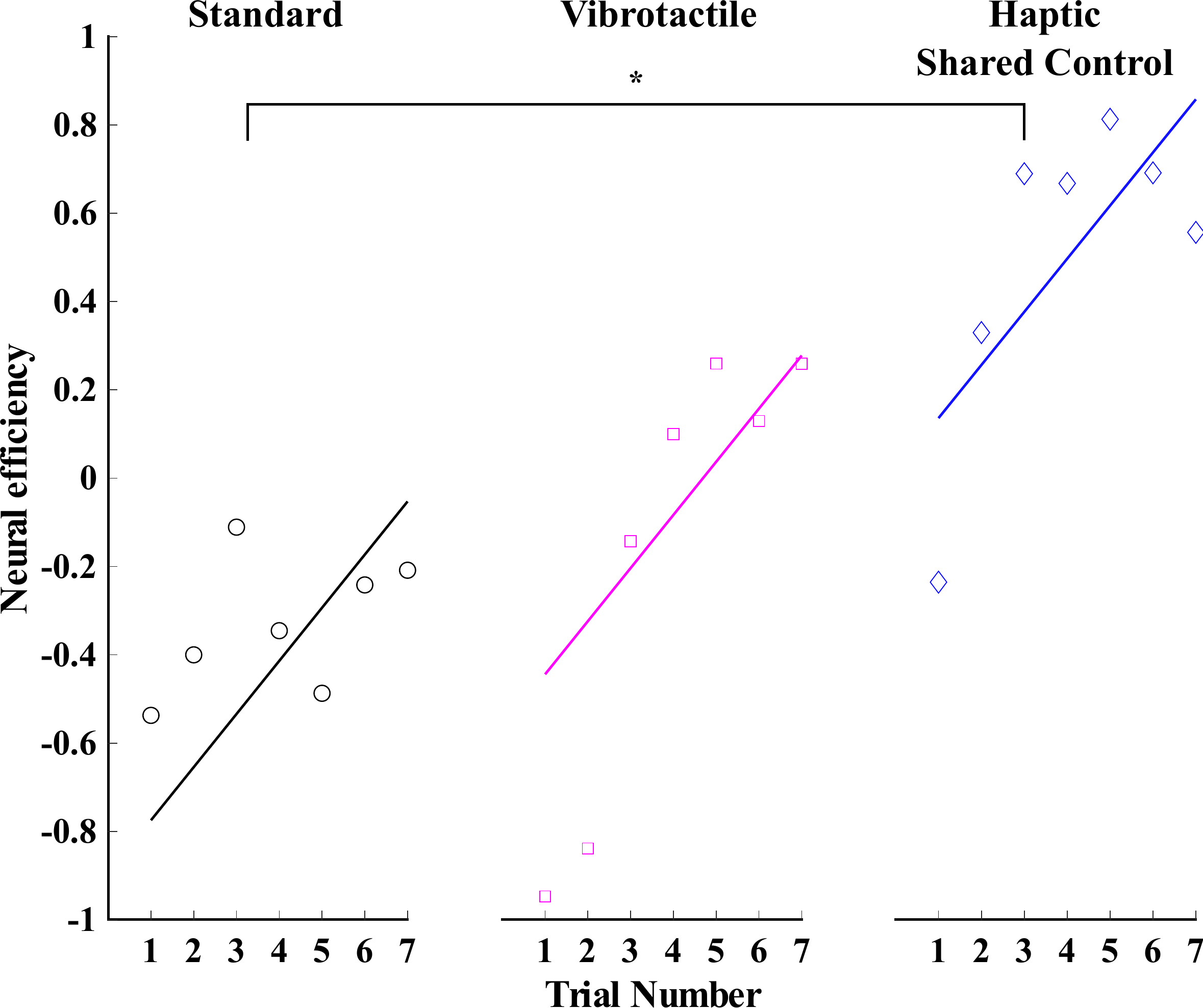} \caption{The neural efficiency for each trial, where the individual data points represent the average for each trial (for all participants in each group), and the solid lines indicate the model's prediction. * indicates $p<0.05$, ** indicates $p<0.01$, and *** indicates $p<0.001$.}
\label{fig: NEHbT4}
\end{figure}

\begin{table*}[!t]
\caption{Summary of model statistics for survey results}
\label{Table:Survey}
\tiny
\centering
\resizebox{\textwidth}{!}{%
\begin{tabular}{lccc|ccc|ccc} 
\toprule

&  \multicolumn{3}{c}{Intercept (Standard)} & \multicolumn{3}{c}{Vibrotactile} &  \multicolumn{3}{c}{Haptic Shared Control}
\\
&  $\beta$ & SE & $p$ & $\beta$ & SE & $p$ & $\beta$ & SE & $p$\\
\midrule
Perceived Performance & 60.8 & 7.05 & $<$ 0.001 & 10.1 & 9.97 & 0.32 & --3.0 & 9.897 & 0.77\\
Mental effort & 44.8 & 3.8 & $<$ 0.001 & 3.8 & 11.6 & 0.75 & 4.8 & 11.6 & 0.68\\
Physical effort & 50.7 & 7.91 & $<$ 0.001 & --9.2 & 11.2 & 0.42 & --4.7 & 11.2 & 0.68\\
Frustration & 30.3 & 7.97 & $<$ 0.001 & --0.6 & 11.3 & 0.96 & 1.5 & 11.3 & 0.89 \\
Time pressure & 37.8 & 9.32 & $<$ 0.001 & 8.6 & 13.2 & 0.53 & --8.6 & 13.2 & 0.52\\
Auditory cues & 64.6 & 9.68 & $<$ 0.001 & 8.5 & 13.7 & 0.54 & --7.1 & 13.7 & 0.61 \\
Visual cues & 81.0 & 5.80 & $<$ 0.001 & --17.4 & 8.20 & 0.04 & 8.2 & 8.20 & 0.33 \\
Haptic cues & 56.1 & 8.80 & $<$ 0.001 & 22.4 & 12.4 & 0.08 & --6.8 & 12.4 & 0.59 \\
\bottomrule
\end{tabular}}
\end{table*}

\subsubsection*{Neural Efficiency}
The neural efficiency indicates the relationship between mental effort and performance. A linear mixed model was used to assess neural efficiency. The neural efficiency in the Standard group was significantly less than zero ($\beta=-0.90, SE=0.30, p=0.005$). The Vibrotactile group was not significantly different from the Standard group ($\beta=0.33, SE=0.37, p=0.38$) or the Haptic Shared Control group ($\beta=--0.58, SE=0.37, p=0.12$). However, the neural efficiency in Haptic Shared Control group was significantly greater than in the Standard group ($\beta=0.91, SE=0.37, p=0.021$). In addition, experience with the task improved neural efficiency overall ($\beta=0.12, SE=0.03, p<0.001$).

\subsection*{Survey}
A linear regression model was used to analyze the survey results. Participants in the Standard group provided ratings for all survey questions that were significantly different from 0 (see Table \ref{Table:Survey} for complete results). The majority of survey responses only significantly differed by group for the following few cases. Participants in the Vibrotactile group rated their use of visual cues as significantly less than the Standard group, and in a post-hoc test with a Bonferroni correction, also less than the Haptic Shared Control group ($\beta=-25.6, SE~=~8.21, p=0.002$). In a post-hoc test with a Bonferroni correction, participants in the Haptic Shared Control group rated their use of somatosensory cues as significantly lower than those in the Vibrotactile group ($\beta$~=~--29.2, $SE$~=~12.45, $p$~=~0.02).





 

\section*{Discussion}

Haptic shared control approaches have been utilized in several human-robot interaction applications with success \cite{Luo2021ADriving,Selvaggio2022ATransportation}; yet, investigations regarding its effectiveness in upper-limb prostheses has been lacking. Furthermore, it is not well understood how a haptic shared control approach affects the human operator's cognitive load and their neural efficiency. To address this gap, we developed a haptic shared control approach for a myoelectric prosthesis and holistically assessed it with both task performance and neurophysiological cognitive load metrics. We compared this control scheme to the standard myoelectric prosthesis and a prosthesis with vibrotactile feedback of grip force in a grasp-and-lift task with a brittle object. The haptic shared control scheme arbitrated between haptically guided control of prosthesis grasping and complete autonomous control of grasping. Here, the autonomous control replicated the human operator's desired grasping strategy in an imitation-learning paradigm.


The primary results indicate that participants in the Haptic Shared Control group exhibited greater neural efficiency -- higher task performance with lower mental effort -- compared to their counterparts in the Standard group. Furthermore, vibrotactile feedback in general was instrumental in appropriately tuning grip force, which is consistent with prior literature \cite{Kim2012, Rosenbaum-Chou2016, Clemente2016Non-InvasiveProstheses}. This benefit combined with the improved dexterity afforded by the autonomous grasp controller substantially improved lifting ability and grip force tuning with the haptic shared control scheme compared to both the Standard and Vibrotactile control schemes. 

Despite the reported benefits of haptic feedback in dexterous task performance and mental effort reduction \cite{Zhou2007,Cao2007CanTime,Thomas2020}, vibrotactile feedback alone was not able to significantly improve lifting ability and neural efficiency compared to Standard control in this study. These results agree with findings from previous investigations on the effect of haptic feedback on grasp-and-lift of a brittle object \cite{Meek1989a, Brown2015a}. This is likely due to the fact that feedback can only inform users of task milestones and task errors after they have occurred. In human sensorimotor control, feedforward control serves to complement feedback strategies by making predictions that guide motor action \cite{Schwartz2016}. 

Thus, for our difficult dexterous task, feedback control strategies alone were insufficient, and had to be supplemented with a feedforward understanding of the appropriate grip force necessary to grasp and lift the fragile object. Once trained, the autonomous controller offloads this burden of myoelectric feedforward control from the user, which results in a marked improvement in both performance and mental effort. This haptic shared control concept leverages the strengths of the human operator's knowledge of the task requirements and subsequently utilizes this experience in tuning the autonomous controller. This finding is in line with the stated benefits of haptic shared control in other human--machine interaction paradigms such as semi-autonomous vehicles and teleoperation \cite{Lazcano2021MPC-BasedDriving, Zhang2021HapticTeleoperation}.

Due to the nature of the haptic shared control scheme in this study, participants in the Haptic Shared Control group  had much less experience with the vibration feedback compared to participants in the Vibrotactile group over the course of the experimental session. Indeed, participants in the Haptic Shared Control group reported significantly less use of somatosensory cues than those in the Vibrotactile group. Thus, it is possible that the boundary between human and machine could become more seamless with additional training on the vibrotactile feedback. Other studies have shown that longer-term and extended training with haptic feedback significantly improved performance \cite{Stepp2012RepeatedPerformance, Clemente2016Non-InvasiveProstheses}. 

It is worth noting here that the auditory cues from the vibrotactile actuator were likely utilized by some participants. Two participants explicitly mentioned that the sound of the tactor was as or even more salient than the tactile sensation itself. Previous research has shown that reaction time decreases with the combination of the tactile and auditory cues from vibrotactile feedback compared to the tactile cues alone \cite{Bao2019VibrotactileTimes}. In addition, it has also been shown that the combination of redundant, multi-modality feedback improves reaction times compared to unimodal feedback \cite{Diederich2004BimodalTime}. This type of incidental feedback is not limited to the audio-tactile cues emanating from the vibrotactor; the sounds generated by the movement of the prosthesis motor were also used by several participants across conditions. Although incidental feedback has been demonstrated to assist in dexterous tasks \cite{Sensinger2020}, it is not enough to achieve the best performance in a task requiring quick and accurate grasp force.


Although we demonstrated the success of the haptic shared control scheme, the present study has some limitations. Only non-amputee participants were evaluated, and the task was conducted with a manufactured object, rather than everyday items. Thus, future work to realize the haptic shared control concept clinically should involve verifying these present results with amputee participants and with a wider range of activities and types of objects, including real-life brittle and fragile objects. A further expansion to the autonomous system includes the ability to recognize object types in order to facilitate switching between different objects and tasks. In addition, the utility of the haptic shared control system should be assessed longitudinally to understand its impact on neural efficiency and direct myoelectric control. 

Existing approaches to shared control within prosthetic systems have focused on supplementing human manual control of the prosthesis with autonomous systems \cite{Zhuang2019SharedProsthesis, Mouchoux2021ImpactProstheses}. These systems do not incorporate haptic feedback, and thus leave the user out of the loop. In contrast, the present study integrates haptic feedback with an autonomous controller in an imitation-learning paradigm, where the autonomous control replicates the desired grasping strategy of the human. Such a system can be expanded and generalized further to facilitate other types of human-robot interaction, such as robotic surgery and human-robot cooperation.

In summary, our results demonstrate that fNIRS can be used to assess cognitive load and neural efficiency in a complex task conducted with a myoelectric prosthesis, and that a haptic shared control system in a myoelectric prosthesis ensures good task performance while incurring low cognitive burden. This is accomplished by the system's individual components (vibrotactile feedback and the imitation-learing controller), whose benefits combine synergistically to optimize performance. These results support the need for hybrid systems in bionic prosthetics to maximize neural and dexterous performance.

\bibliography{references}



\section*{Acknowledgments}
We would like to thank Garrett Ung for designing and constructing the instrumented object. Thanks also to Leah Jager for providing her statistical consulting services.
We also thank the National Science Foundation for funding the first author through a Graduate Research Fellowship. 

\section*{Author contributions statement}
NT conceived the experimental design and integrated the hardware and software to run the experiment. NT and AM ran the user study. NT analyzed the data. HA advised on the experimental design, use of fNIRS and statistical analyses, and JDB advised on the general experiment and statistical analyses. NT wrote the manuscript draft. All authors edited the manuscript and approve its submission. 

\section*{Data Availability}
The datasets generated during and/or analyzed during the current study are available from the corresponding author on reasonable request.

\section*{Additional information}

\textbf{Competing interests} fNIR Devices, LLC manufactures the optical brain imaging instrument and licensed IP and know-how from Drexel University. Dr. Ayaz was involved in the technology development and thus offered a minor share in the startup firm fNIR Devices. The authors declare that the research was conducted in the absence of any commercial or financial relationships that could be construed as a potential conflict of interest.





\end{document}